\documentclass[sigconf]{acmart}

\AtBeginDocument{%
  \providecommand\BibTeX{{%
    \normalfont B\kern-0.5em{\scshape i\kern-0.25em b}\kern-0.8em\TeX}}}

\setcopyright{rightsretained}
\acmConference[KDD Converse'20]{KDD Workshop on Conversational Systems Towards Mainstream Adoption}{August 2020}{}
\acmYear{2020}
\copyrightyear{2020}
\makeatletter
\renewcommand\@formatdoi[1]{\ignorespaces}
\makeatother
\acmISBN{}

\begin{document}

\title{Abstractive Summarization of \\ Spoken and Written Instructions with BERT}

\author{Alexandra Savelieva}
\authornote{Equal contribution.}
\authornote{Also with Microsoft.}
\affiliation{%
  \institution{UC Berkeley School of Information}
}
\email{saveale@ischool.berkeley.edu}

\author{  Bryan Au-Yeung}
\authornotemark[1]
\affiliation{%
  \institution{UC Berkeley School of Information}
}
\email{bkauyeung@berkeley.edu}

\author{Vasanth Ramani}
\authornotemark[1]
\authornotemark[2]
\affiliation{%
  \institution{UC Berkeley School of Information}
  }
  
\email{rlvasanth@ischool.berkeley.edu }

\renewcommand{\shortauthors}{Savelieva, Au-Yeung, Ramani}

\begin{abstract}

Summarization of speech is a difficult problem due to the spontaneity of the flow, disfluencies, and other issues that are not usually encountered in written texts. Our work presents the first application of the BERTSum model to conversational language. We generate abstractive summaries of narrated instructional videos across a wide variety of topics, from gardening and cooking to software configuration and sports. In order to enrich the vocabulary, we use transfer learning and pretrain the model on a few large cross-domain datasets in both written and spoken English. We also do preprocessing of transcripts to restore sentence segmentation and punctuation in the output of an ASR system. The results are evaluated with ROUGE and Content-F1 scoring for the How2 and WikiHow datasets. We engage human judges to score a set of summaries randomly selected from a dataset curated from  HowTo100M and YouTube. Based on blind evaluation, we achieve a level of textual fluency and utility close to that of summaries written by human content creators. The model beats current SOTA when applied to WikiHow articles that vary widely in style and topic, while showing no performance regression on the canonical CNN/DailyMail dataset.  Due to the high generalizability of the model across different styles and domains, it has great potential to improve accessibility and discoverability of internet content. We envision this integrated as a feature in intelligent virtual assistants, enabling them to summarize both written and spoken instructional content upon request.

\end{abstract}



\begin{teaserfigure}
  \includegraphics[width=\textwidth]{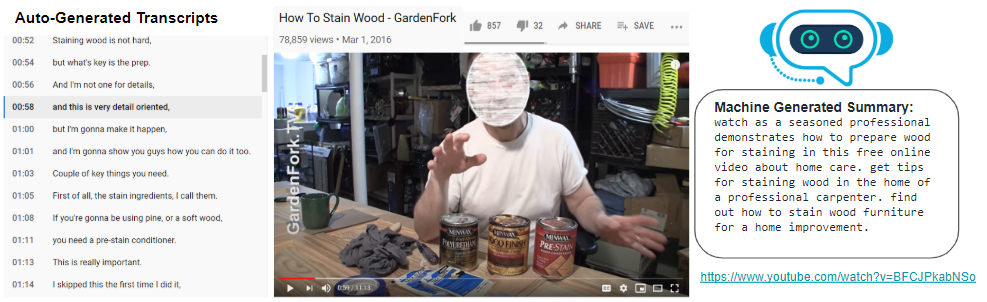}
  \caption{A screenshot of a How2 YouTube video with transcript and model generated summary.}
  \Description{Enjoying the baseball game from the third-base
  seats. Ichiro Suzuki preparing to bat.}
  \label{fig:teaser}
\end{teaserfigure}

\maketitle
\section{Introduction}

\begin{figure*}
\centering
  \includegraphics[width=\textwidth]{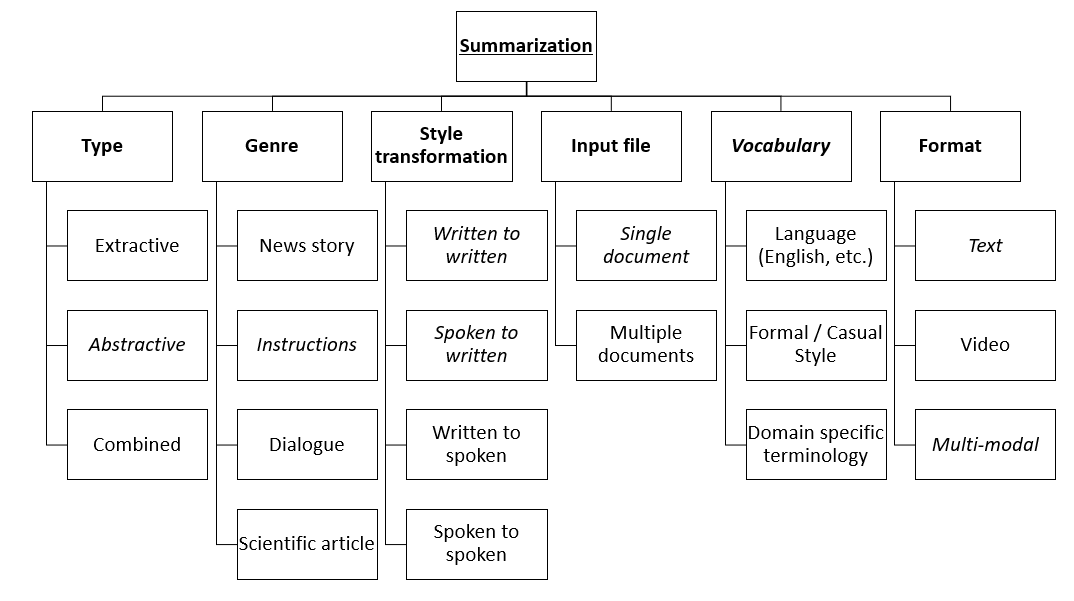}
  \caption{A taxonomy of summarization types and methods.}
  \Description{Enjoying the baseball game from the third-base
  seats. Ichiro Suzuki preparing to bat.}
  \label{fig:taxonomy}
\end{figure*}

The motivation behind our work involves making the growing amount of user-generated online content more accessible. In order to help  users digest information, our research focuses on improving automatic summarization tools. Many creators of online content use a variety of casual language, filler words, and professional jargon. Hence, summarization of text implies not only an extraction of important information from the source, but also a transformation to a more coherent and structured output. In this paper we focus on both extractive and abstractive summarization of narrated instructions in both written and spoken forms. Extractive summarization is a simple classification problem for identifying the most important sentences in the document and classifies whether a sentence should be included in the summary. Abstractive summarization, on the other hand, requires language generation capabilities to create summaries containing novel words and phrases not found in the source text. Language models for summarization of conversational texts often face issues with fluency, intelligibility, and repetition. This is the first attempt to use BERT-based  model for summarizing  spoken language from ASR (speech-to-text) inputs.  We are aiming to develop a generalized tool that can be used across a variety of domains for How2 articles and videos. Success in solving this problem opens up possibilities for extension of the summarization model to other applications in this area, such as summarization of dialogues in conversational systems between humans and bots \cite{liu2019}. 

The rest of this paper is divided in the following sections:
\begin{itemize}
\item A review of state-of-the art summarization methods;
\item A description of dataset of texts, conversations, and summaries used for training;
\item Our application of BERT-based text summarization models \cite{inproceedings} and fine tuning on auto-generated scripts from instructional videos; 
\item Suggested improvements to evaluation methods in addition to the metrics \cite{10.3115/1118162.1118168} used by previous research.
\item Analysis of experimental results and comparison to benchmark
\end{itemize}

\section{Prior work}
A taxonomy of summarization types and methods is presented in Figure \ref{fig:taxonomy}.  Prior to 2014, summarization was centered on extracting lines from single documents using statistical models and neural networks with limited success \cite{DBLP:conf/emnlp/SvoreVB07} \cite{inproceedings}. The work on sequence to sequence models from Sutskever et al. \cite{DBLP:journals/corr/SutskeverVL14} and Cho et al. \cite{DBLP:journals/corr/ChoMGBSB14} opened up new possibilities for neural networks in natural language processing. From 2014 to 2015, LSTMs (a variety of RNN) became the dominant approach in the industry which achieved state of the art results. Such architectural changes became successful in tasks such as speech recognition, machine translation, parsing, image captioning. The results of this paved the way for abstractive summarization, which began to score competitively against extractive summarization. In 2017, a paper by Vaswani et.al.  \cite{DBLP:journals/corr/VaswaniSPUJGKP17} provided a solution to the ‘fixed length vector’ problem, enabling neural networks to focus on important parts of the input for prediction tasks. Applying attention mechanisms with transformers became more dominant for tasks, such as translation and summarization.

In abstractive video summarization, models which incorporate variations of LSTM and deep layered neural networks have become state of the art performers. In addition to textual inputs, recent research in multi-modal summarization incorporates visual and audio modalities into language models to generate summaries of video content. However, generating compelling summaries from conversational texts using transcripts or a combination of modalities is still challenging. The deficiency of human annotated data has limited the amount of benchmarked datasets available for such research \cite{palaskar-etal-2019-multimodal} \cite{li-etal-2017-multi}. Most work in the field of document summarization relies on structured news articles. Video summarization focuses on heavily curated datasets with structured time frames, topics, and styles \cite{1221239}. Additionally, video summarization has been traditionally accomplished by isolating and concatenating important video frames using natural language processing techniques \cite{DBLP:conf/icmcs/ErolLH03}. Above all, there are often inconsistencies and stylistic changes in spoken language that are difficult to translate into written text. In this work, we approach video summarizations by extending top performing single-document text summarization models \cite{rush-etal-2015-neural} to a combination of narrated instructional videos, texts, and news documents of various styles, lengths, and literary attributes.

\section{Methodology}

\subsection{Data Collection}
We hypothesize that our model's ability to form coherent summaries across various texts will benefit from training across larger amounts of data. Table \ref{tab:datasets} illustrates various textual and video dataset sizes. All training datasets include written summaries. The language and length of the data span from informal to formal and single sentence to short paragraph styles.

\begin{table}
  \caption{Training and Testing Datasets}
  \label{tab:datasets}
  \centering
  \begin{tabular}{llll}
    \toprule
        \textbf{Total Training Dataset Size} & \textbf{535,527} \\
    \midrule
    CNN/DailyMail & 90,266 and 196,961 \\
    \midrule
    WikiHow Text & 180,110 \\
    \midrule
    How2 Videos & 68,190 \\
    \midrule
  \textbf{Total Testing Dataset Size}  &  \textbf{5,195 videos} \\
    \midrule
    YouTube (DIY Videos and How-To Videos)& 1,809 \\
    \midrule
    HowTo100M & 3,386   \\
   \bottomrule
  \end{tabular}
  \end{table}

\begin{itemize}

\item \textbf{CNN/DailyMail dataset} \cite{NIPS2015_5945}: CNN and DailyMail includes a combination of news articles and story highlights written with an average length of 119 words per article and 83 words per summary. Articles were collected from 2007 to 2015. 

\item \textbf{Wikihow dataset} \cite{DBLP:journals/corr/abs-1810-09305}: a large scale text dataset containing over 200,000 single document summaries. Wikihow is a consolidated set of recent ‘How To’ instructional texts compiled from wikihow.com, ranging from topics such as ‘How to deal with coronavirus anxiety’ to ‘How to play Uno.’ These articles vary in size and topic but are structured to instruct the user. The first sentences of each paragraph within the article are concatenated to form a summary. 

\item \textbf{How2 Dataset} \cite{DBLP:journals/corr/abs-1811-00347}:  This YouTube compilation has videos (8,000 videos - approximately 2,000 hours) averaging 90 seconds long and 291 words in transcript length. It includes human written summaries which video owners were instructed to write summaries to maximize the audience. Summaries are two to three sentences in length with an average length of 33 words. 

\end{itemize}

Despite the development of instructional datasets such as Wikihow and How2, advancements in summarization have been limited by the availability of human annotated transcripts and summaries. Such datasets are difficult to obtain and expensive to create, often resulting in repetitive usage of singular-tasked and highly structured data . As seen with samples in the How2 dataset, only the videos with a certain length and structured summary are used for training and testing. To extend our research boundaries, we complemented existing labeled summarization datasets with auto-generated instructional video scripts and human-curated descriptions. 

We introduce a new dataset obtained from combining several 'How-To' and Do-It-Yourself YouTube playlists along with samples from the published HowTo100Million Dataset \cite{DBLP:journals/corr/abs-1906-03327}. To test the plausibility of using this model in the wild, we selected videos across different conversational texts that have no corresponding summaries or human annotations. The selected 'How-To' \cite{How-toVideos}) and 'DIY'\cite{DIYHow-toVideos} datasets are instructional playlists covering different topics from downloading mainstream software to home improvement. The 'How-To' playlist uses machine voice-overs in the videos to aid instruction while the 'DIY' playlist has videos with a human presenter. The HowTo100Million Dataset is a large scale dataset of over 100 million video clips taken from narrated instructional videos across 140 categories. Our dataset incorporates a sample across all categories and utilizes the natural language annotations from automatically transcribed narrations provided by YouTube.

\begin{table}
  \caption{Additional Dataset Statistics}
  \label{tab:datasets2}
  \centering
  \begin{tabular}{llll}
  \toprule
   YouTube Min / Max Length  &  4 / 1,940 words     \\
\midrule
YouTube Avg Length & 259 words    \\
\midrule
  HowTo100M Sample Min / Max Length & 5 / 6,587 words    \\
\midrule
HowTo100M Sample Avg Length & 859 words   \\
\bottomrule
  \end{tabular}
\end{table}

\subsection{Preprocessing}
\label{Preprocessing}

\begin{figure*}
\centering
  \includegraphics[width=\textwidth]{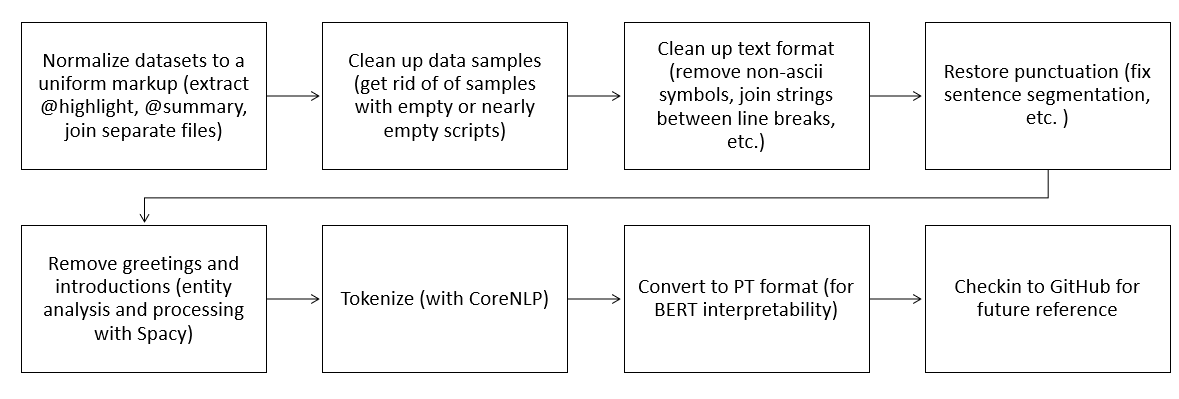}
  \caption{A pipeline for preprocessing of texts for summarization.}
  \Description{Enjoying the baseball game from the third-base
  seats. Ichiro Suzuki preparing to bat.}
  \label{fig:preprocessing}
\end{figure*}

Due to diversity and complexity of our input data, we built a preprocessing pipeline for aligning the data to a common format. We observed issues with lack of punctuation, incorrect wording, and extraneous introductions which impacted model training. With these challenges, our model misinterpreted text segment boundaries and produces poor quality summaries. In exceptional cases, the model failed to produce any summary. In order to maintain the fluency and coherency in human written summaries, we cleaned and restored sentence structure as shown in the Figure  \ref{fig:preprocessing}. We applied  entity detection from an open-source software library for advanced natural language processing called  \verb+spacy+ \cite{spacy2} and \verb+nltk+: the Natural Language Toolkit for symbolic and statistical natural language processing \cite{journals/corr/cs-CL-0205028} to remove introductions and anonymize the inputs of our summarization model. We split sentences and tokenized using the Stanford Core NLP toolkit on all datasets and preprocessed the data in the same method used by See et.al. in  \cite{DBLP:journals/corr/SeeLM17}.

\subsection{Summarization models}

We utilized the BertSum models proposed in \cite{liu-lapata-2019-text}  for our research. This includes both Extractive and Abstractive summarization models, which employs a document level encoder based on Bert. The transformer architecture applies a pretrained BERT encoder with a randomly initialized Transformer decoder. It uses two different learning rates: a low rate for the encoder and a separate higher rate for the decoder to enhance learning.

We used a 4-GPU Linux machine and initialized a baseline by training an extractive model on 5,000 video samples from the How2 dataset.  Initially, we applied BERT base uncased with 10,000 steps and fine tuned the summarization model and BERT layer, selecting the top-performing epoch sizes. We followed this initial model by training the abstractive model on How2 and WikiHow individually.

The best version of the abstractive summarization model was trained on our aggregated dataset of CNN/DailyMail, Wikihow, and How2 datasets with a total of 535,527 examples and 210,000 steps. We used a training batch size of 50 and ran the model for 20 epochs. By controlling the order of datasets in which we trained our model, we were able to improve the fluency of summaries. As stated in previous research, the original model contained more than 180 million parameters and used two Adam optimizers with $\beta_1$=0.9 and $\beta_2$ =0.999 for the encoder and decoder respectively. The encoder used a learning rate of 0.002 and the decoder had a learning rate of 0.2 to ensure that the encoder was trained with more accurate gradients while the decoder became stable. The results of experiments are discussed in Section \ref{Experiments}.

We hypothesized that the training order is important to the model in the same way humans learn. The idea of applying \emph{curriculum learning} \cite{conf/icml/BengioLCW09} in natural language processing has been a growing topic of interest \cite{xu-etal-2020-curriculum}. We begin training on highly structured samples before moving to more complicated, but predictable language structure \ref{Experiments}. Only after training textual scripts do we proceed to video scripts, which presents additional challenges of ad-hoc flow and conversational language.

\subsection{Scoring of results}
\label{Scoring}
Results were scored using ROUGE, the standard metric for abstractive summarization {\cite{lin-2004-rouge}}. While we expected a correlation between good summaries and high ROUGE scores, we observed examples of poor summaries with high scores, such as in Figure \ref{fig:funnysummary}, and good summaries with low ROUGE scores.  Illustrative example of why ROUGE metrics is not sufficient is presented in Appendix, Figure \ref{fig:funnysummary}.

Additionally, we added Content F1 scoring, a metric proposed by Carnegie Mellon University {\cite{denkowski:lavie:meteor-wmt:2014}} to focus on the relevance of content. Similar to ROUGE, Content F1 scores summaries with a weighted f-score and a penalty for incorrect word order. It also discounts stop and buzz words that frequently occur in the How-To domain, such as “learn from experts how to in this free online video”.  

To score passages with no written summaries, we surveyed human judges with a framework for evaluation using Python, Google Forms, and Excel spreadsheets. Summaries included in the surveys are randomly sampled from our dataset to avoid biases. In order to avoid asymmetrical information between human versus machine generated summaries, we removed capitalized text. We asked two types of questions: A Turing test question for participants to distinguish AI from human-generated descriptions. The second involves selecting quality ratings for summaries. Below are definitions of criteria for clarity:
\begin{itemize}

\item Fluency: Does the text have a natural flow and rhythm?
\item Usefulness: Does it have enough information to make a user decide whether they want to spend time watching the video?
\item Succinctness: Does the text look concise or do does it have redundancy?
\item Consistency: Are there any non sequiturs - ambiguous, confusing or contradicting statements in the text?
\item Realisticity: Is there anything that seems far-fetched and bizarre in words combinations and doesn't look "normal"?

\end{itemize}

Options for grading summaries are as follows: 1: Bad   2: Below Average   3: Average   4: Good   5: Great.

\section{Experiments and Results}
\label{Experiments}
\subsection{Training}

BertSum model is the best performing model on the CNN/DailyMail dataset producing state-of-the-art results (Row 6) \ref{table1}. BertSum model supports both extractive and abstractive summarization techniques. Our baseline results were obtained from applying this extractive BertSum model pretrained on CNN/DailyMail to How2 videos. But the model produced very low scores for our scenario. Summaries generated from the model were incoherent, repetitive, and uninformative. Despite poor performance, the model performed better in the health sub-domain within How2 videos. We explained this as a symptom of heavy coverage in news reports generated by CNN/DailyMail. We realized that extractive summarization is not the strongest model for our goal: most YouTube videos are presented with a casual conversational style, while summaries have higher formality. We pivoted to abstractive summarization to improve performance. 

Abstractive model uses an encoder-decoder architecture, combining the same pretrained BERT encoder with a randomly initialized Transformer decoder. It uses a special technique where the encoder portion is almost kept same with a very low learning rate and creates a separate learning rate for the decoder to make it learn better. In order to create a generalizable abstractive model, we first trained on a large corpus of news. This allowed our model to understand structured texts. We then introduced Wikihow, which exposes the model to the How-To domain. Finally, we trained and validated on the How2 dataset, narrowing the focus of the model to a selectively structured format. In addition to ordered training, we experimented with training the model using random sets of homogeneous samples. We discovered that training the model using an ordered set of samples performed better than random ones.

\begin{figure}
  \centering
  \includegraphics[width=\linewidth]{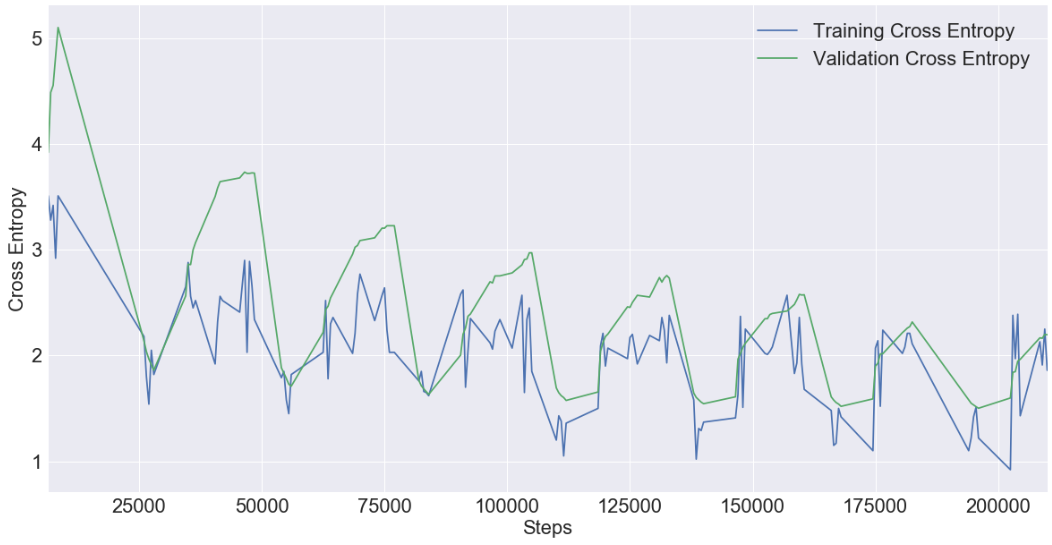}
  \caption{Cross Entropy: Training vs Validation}
  \label{fig:xent}
\end{figure}%

\begin{figure} 
  \centering
  \includegraphics[width=\linewidth]{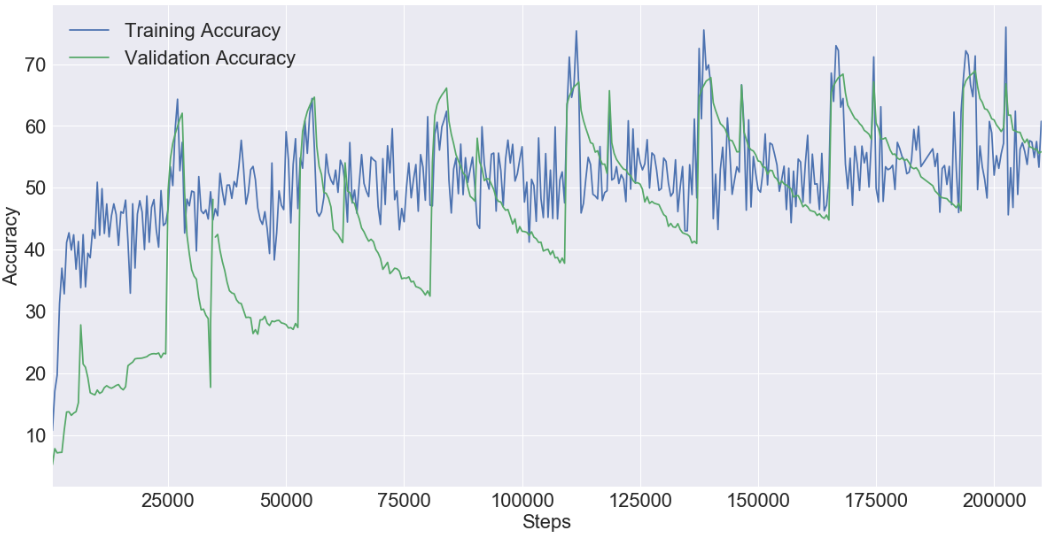}
  \caption{Accuracy: Training vs Validation}
  \label{fig:accuracy}
\end{figure} 

The cross entropy chart  in the Figure \ref{fig:xent} shows that the model is neither overfitting nor underfitting the training data. Good fit is indicated with the convergence of training and validation lines. Figure \ref{fig:accuracy} shows the model’s accuracy metric on the training and validation sets. The model is validated using the How2 dataset against the training dataset. The model improves as expected with more steps.

\subsection{Evaluation}

The BertSum model trained on CNN/DailyMail \cite{liu-lapata-2019-text}  resulted in state of the art scores when applied to samples from those datasets. However, when tested on our How2 Test dataset, it gave very poor performance and a lack of generalization in the model (see Row 1 in Table~\ref{table1}). Looking at the data, we found that the model tends to pick the first one or two sentences for the summary.  We hypothesized that removing introductions from the text would help improve ROUGE scores. Our model improved a few ROUGE points after applying  preprocessing described in the Section \ref{Preprocessing} above. Another improvement came from adding word deduping to the output of the model, as we observed it occurring on rare words which are unfamiliar to the model. We still did not achieve scores get higher than 22.5 ROUGE-1 F1 and 20 ROUGE-L F1 (initial scores achieved from training with only the CNN/DailyMail dataset and tested on How2 data). Reviewing scores and texts of individual summaries showed that the model performed better on some topics such as medicine, while scoring lower on others, such as sports. 

The differences in conversational style of the video scripts and news stories (on which the models were pretrained) impacted the quality of the model output. In our initial application of the extractive summarization model pretrained on CNN/DailyMail dataset, stylistic errors manifested in a distinct way. The model considered initial introductory sentences to be important in generating summaries (this phenomena is referred to by [15] as N-lead, where N is the number of important first sentences). Our model generated short, simple worded summaries such as "hi!" and "hello, this is <first and last name>".

Retraining abstractive BertSum on How2 gave a very interesting unexpected result - the model converged to a state of spitting out the same meaningless summary of buzzwords that are common for most videos, regardless of the domain: \emph{"learn how to do the the the a in this free video clip clip clip series clip clip on how to make a and expert chef and expert in this unique and expert and expert. to utilize and professional . this unique expert for a professional."}

In our next series of experiments, we used extended dataset for training.  Even though the difference in ROUGE scores for the results from BertSum Model 1 (see Table~\ref{table1}) are not drastically different from  BertSum Models 2 and 3, the quality of summaries from the perspective of human judges is qualitatively different.

Our best results on How2 videos (see experiment 4 in Table~\ref{table1}) were accomplished by leveraging the full set of labeled datasets (CNN/DM, WikiHow, and How2 videos) with order preserving configuration. The best ROUGE scores we got for video summarization are comparable to best results among news documents \cite{liu-lapata-2019-text} (see row 9 in Table~\ref{table1}).

Finally, we beat current best results on WikiHow. The current benchmark Rouge-L score for WikiHow dataset is 26.54 in Row 8. \ref{table1} Our model uses the BERT abstractive summarization model to produce a Rouge-L score of 36.8 in Row 5 \ref{table1}, outperforming the current benchmark score by 10.26 points. Compared to  Pointer Generator+Coverage model, the improvement on Rouge-1 and Rouge-L is about 10 points each. We got same results testing for WikiHow using BertSum with ordered training on the the entire How2, WikiHow, and CNN/DailyMail dataset.

\begin{table*}
\caption{Comparison of results}
\label{table1}
\centering
\begin{tabular}{llllll}
\toprule
\multicolumn{2}{c}{Experiment} \\
\cmidrule(r){1-2}
Model & Pretraining Data & Test Set & Rouge-1 &Rouge-L &Content-F1\\
\midrule
1. BertSum, BertSum with pre+post processing & CNN/DM & How2 & 18.08 to 22.47 & 18.01 to 20.07 & 26.0 \\
\midrule
2. BertSum with random training & How2, 1/50 Sampled- & How2 & 24.4 &21.45 & 18.7 \\
& WikiHow, CNN/DM &\\
\midrule
3. BertSum with random training and & How2, 1/50 Sampled- & How2 & 26.32 &22.47 & 32.9 \\
postprocessing & WikiHow, CNN/DM &\\
\midrule
4. BertSum with ordered training & How2, WikiHow, & How2 & 48.26 &44.02 & 36.4 \\
  & CNN/DM &\\
\midrule
5. BertSum & WikiHow & WikiHow & 35.91 & 34.82 & 29.8 \\
\midrule
\midrule

6. BertSum \cite{liu-lapata-2019-text} & CNN/DM & CNN/DM & 43.23 &39.63 & Out of Scope \\
\midrule
7. Multi-modal Model \cite{palaskar-etal-2019-multimodal} & How2 & How2 & 59.3 &59.2 & 48.9 \\
\midrule
8. MatchSum (BERT-base) \cite{Zhong2020ExtractiveSA} & WikiHow & WikiHow & 31.85 &29.58 & Not Available \\
\midrule
9. Lead 3 for WikiHow \cite{DBLP:journals/corr/abs-1810-09305} & Not Applicable & CNN/DM & 40.34 &36.57 & Not Available \\
\midrule
10. Lead 3 for CNN/DM \cite{DBLP:journals/corr/abs-1810-09305} & Not Applicable & WikiHow & 26.00 &24.25 & Not Available \\
\midrule
11. Lead 3 for How2 \cite{DBLP:journals/corr/abs-1810-09305}& Not Applicable & How2 & 23.66 & 20.69 & 16.2 \\
\bottomrule
\end{tabular}
\end{table*}

With our initial  results, we achieved fluent and understandable video descriptions which give a clear idea about the content. Our scores did not surpass scores from other researchers \cite{DBLP:journals/corr/abs-1811-00347} despite employing BERT. However, our summaries appear to be more fluent and useful in content for users looking at summaries in the How-To domain. Some examples are given in the [Appendix: \ref{AppendixC}].

Abstractive summarization was helpful for reducing the effects of speech-to-text errors, which we observed in some videos transcripts, especially auto-generated closed captioning in the additional dataset that we created as part of this project (transcripts in How2 videos were reviewed and manually corrected, so spelling errors there are less common).  For example, in one of the samples in our test dataset closed captioning confuses the speaker’s words “how you get a text from a YouTube video” for  “how you get attacks from a YouTube video”. As there is usually a lot of redundancy in explanations, the model is still able to figure out sufficient context to produce a meaningful summary. We did not observe situations where the summaries did not match topic of the video due to impact from spelling errors that frequently occur in ASR-generated scripts without human supervision, but ensuring correct boundaries between sentences by using Spacy to fix punctuation errors at preprocessing stage made a very big difference.

\begin{figure} 
  \centering
  \includegraphics[width=\linewidth]{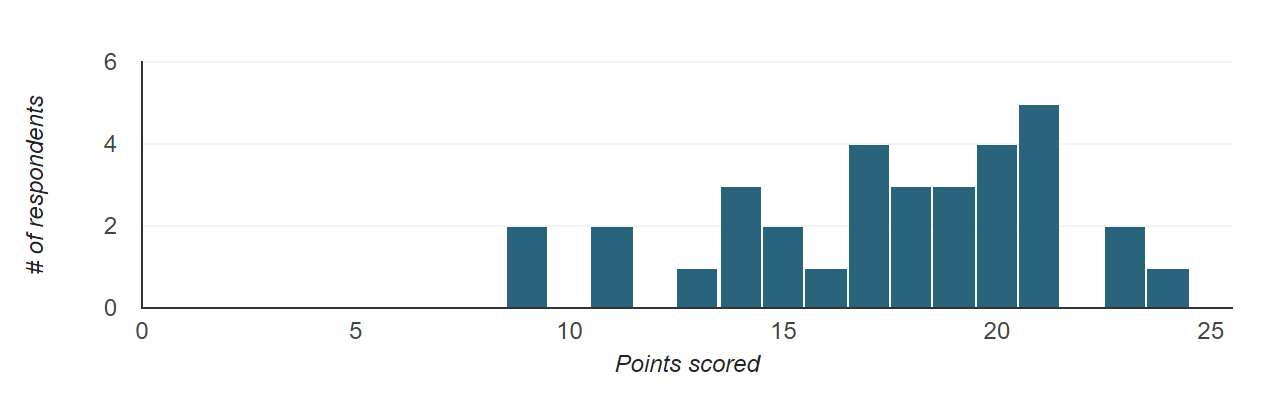}
  \caption{Scores of human judges in the challenge to  distinguishing ML-generated summaries from actual video annotations on YouTube }
  \label{fig:scores}
\end{figure}

\begin{figure} 
  \centering
  \includegraphics[width=\linewidth]{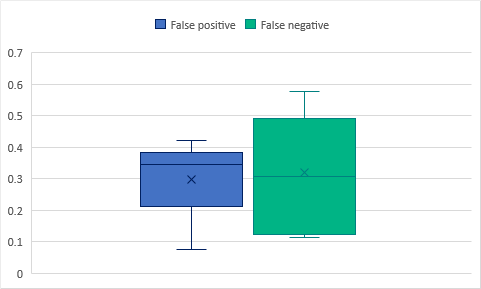}
  \caption{Distribution of average FP and FN ratio per question}
  \label{fig:box}
\end{figure}

\begin{figure} 
  \centering
  \includegraphics[width=\linewidth]{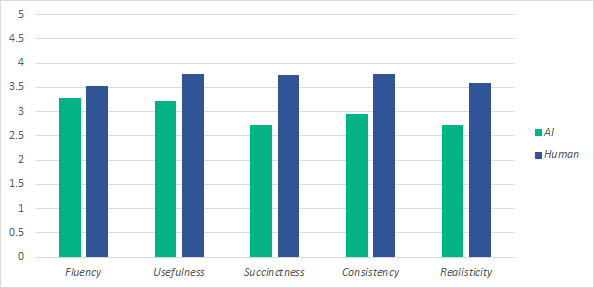}
  \caption{Quality assessment of generated summaries}
  \label{fig:quality}
\end{figure}

Based on these observations, we decided that the model generated strong results comparable to human written descriptions. To analyze the differences in summary quality, we leveraged help of human experts to evaluate conversational characteristics between our summaries and the descriptions that users provide for their videos on YouTube. We recruited a diverse group of 30+ volunteers to blindly evaluate a set of 25 randomly selected video summaries that were generated by our model and video descriptions from our curated conversational dataset. We created two types of questions: one, a version of famous Turing test, was a challenge to distinguish AI from human-curated descriptions and used the framework described in Section \ref{Scoring}. Participants were made aware that there is equal possibility that some, all, or none of these summary outputs were machine generated in this classification task. The second question collects a distribution of ratings addressing conversation quality. The aggregated results for both evaluations are in Figures \ref{fig:scores} - \ref{fig:quality}. We observe zero perfect scores on Turing test answers. Results included many false positives and false negatives [Appendix: \ref{AppendixD}]. 

The quality of our test output is comparable to YouTube summaries. "Realistic" text is the main growth opportunity because the abstractive model is prone to generating incoherent sentences that are grammatically correct. Human authors are prone to making language use errors. The advantage of using abstractive summarization models allows us to mitigate some issues with video author's grammar. 


\section{Conclusion}

The contributions of our research address multiple issues that we identified in pursuit of generalizing BertSum model for summarization of instructional video scripts throughout the training process. 

\begin{itemize}

\item We explored how different combinations of training data and parameters  impact the training performance  of BertSum abstractive summarization model.
\item We came up with novel preprocessing steps for auto-generated closed captioning scripts before summarization.
\item We generalized BertSum abstractive summarization model to auto-generated instructional video scripts with the quality level that is close to randomly sampled descriptions created by YouTube users.
\item We designed and implemented a new framework for blind unbiased review that produces more actionable and objective scores,  augmenting ROUGE, BLEU and Content F1.
\end{itemize}
 
All the artifacts listed above are available  in to our repository for the benefit of future researchers \footnote{https://github.com/alebryvas/berk266/ - it's not a public repository yet, but we can provide access upon request}. Overall, the results we obtained by now on amateur narrated instructional videos  make us believe that we were able to come up with a trained model  that generates summaries from ASR (speech-to-text) scripts of competitive quality to human-curated descriptions on YouTube. With the limited availability of labeled summary datasets, our future plan is to create several benchmark models to extend the human valuation framework with human curated summaries. Given the successes of generalized summaries across informal and formal styles of conversation, we believe that investigating the application of these summarization models to human - chatbot dialogues is an important direction for future work.

\begin{acks}
We would first like to thank Shruti Palaskar, whose mentorship and guidance were invaluable throughout our research. We would also like to thank Jack Xue, James McCaffrey, Jiun-Hung Chen, Jon Rosenberg, Isidro Hegouaburu, Sid Reddy, Mike Tamir, and Troy Deck for their insights and feedback. We also thank the survey participants for taking time to complete our human evaluation study. 
\end{acks}

\bibliographystyle{ACM-Reference-Format}
\bibliography{sample-base}

\appendix

\section{Model details}
Extractive summarization is generally a binary classification task with labels indicating whether sentences should be included in the summary. Abstractive summarization, on the other hand, requires language generation capabilities to create summaries containing novel words and phrases not found in the source text. 

\begin{figure}
  \centering
  \includegraphics[scale=0.5]{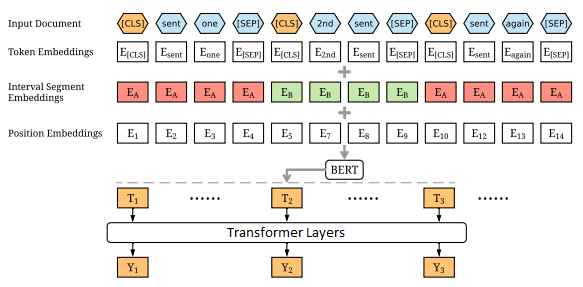}
  \caption{BertSum Architecture. From \cite{liu-lapata-2019-text} }
  \label{fig:architecure}
\end{figure}

The architecture in the Figure \ref{fig:architecure} shows the BertSum model. It uses a novel documentation level encoder based on BERT which can encode a document and obtain representation for the sentences. CLS token is added to every sentence instead of just 1 CLS token in the original BERT model. Abstractive model uses an encoder-decoder architecture, combining the same pretrained BERT encoder with a randomly initialized Transformer decoder. The model uses a special technique where the encoder portion is almost kept same with a very low learning rate and a separate learning rate is used for the decoder to make it learn better. 

\section{An Illustrative Example of why ROUGE metrics are not sufficient}

\begin{figure}[H]
  \includegraphics[width=\linewidth]{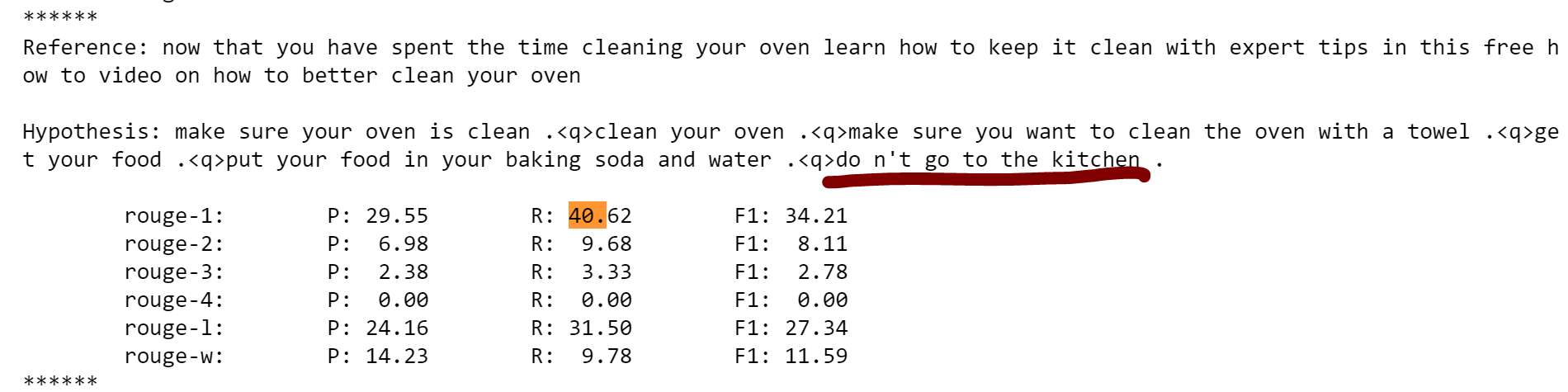}
  \caption{An example where ROUGE metric is confusing}
  \label{fig:funnysummary}
\end{figure}

\section{Examples of Comparison of our model output vs Benchmark \cite{palaskar-etal-2019-multimodal} and reference summaries}
\label{AppendixC}
Below examples were selected to illustrate several aspects of the problem. First, we share URLs of the videos so that the reader may view the original content. Second, we share the final result of abstractive summarization with our current best model version (Summary Abs). For comparison, we provide summaries from current Benchmark for How2 videos that bypasses our model in terms of scores, but, as can be seen in these examples, not in the fluency and usefulness. Reference represents the actual YouTube video description curated by the authors. For contrast, we show Summary Ext - the result of extractive summarization, which explains why abstractive summarization is a better fit for the purpose, as we are trying to accomplish style conversion from spoken for the source text to written for the target summary. Since BertSum is uncased, all texts below were converted to lower case for consistency. 
\begin{itemize}

\item Video 1: \url{https://www.youtube.com/watch?v=F_4UZ3bGMP8}
\item Summary Abs 1: growing rudbeckia requires full hot sun and good drainage. grow rudbeckia with tips from a gardening specialist in this free video on plant and flower care. care for rudbeckia with gardening tips from an experienced gardener.
\item Benchmark 1: growing black - eyed - susan is easy with these tips, get expert gardening tips in this free gardening video .
\item Reference 1: growing rudbeckia plants requires a good deal of hot sun and plenty of good drainage for water. start a rudbeckia plant in the winter or anytime of year with advice from a gardening specialist in this free video on plant and flower care. 
\item Summary Ext 1: make sure that your plants are in your garden. get your plants. don't go to the flowers. go to your garden's soil. put them in your plants in the water. take care of your flowers. 
\item Video 2: \url{https://www.youtube.com/watch?v=LbsGHj2Akao}
\item Summary 2: camouflage thick arms by wearing sleeves that are not close to the arms and that have a line that goes all the way to the waist. avoid wearing jackets and jackets with tips from an image consultant in this free video on fashion. learn how to dress for fashion modeling.
\item Benchmark 2: hide thick arms and arms by wearing clothes that hold the arms in the top of the arm. avoid damaging the arm and avoid damaging the arms with tips from an image consultant in this free video on fashion.
\item Reference 2: hide thick arms by wearing clothes sleeves that almost reach the waist to camouflage the area .conceal the thickness at the top of the arms with tips from an image consultant in this free video on fashion.
\item Summary Ext 2: make sure that you have a look at the top of the top. if you want to wear the right arm. go to the shoulder. wear a long-term shirts. keep your arm in your shoulders. don't go out.

\section{Examples of False Positives and False Negatives from Survey Results}
\label{AppendixD}
False Negative (FN): Survey participants believed sample summaries were written by robots when sample were written by humans.  

False Positive (FP): Survey participants believed sample summaries were written by humans when sample were written by robot.

FN Examples: 

\item "permanently fix flat atv tires with tireject ??. dry rot, bead leaks, nails, sidewall punctures are no issue. these 30yr old atv tires permanently sealed and back into service in under 5 min. they sealed right up and held air for the first time in a long time. this liquid rubber and kevlar are a permanent repair and will protect from future punctures."

\item "how to repair a bicycle tire : how to remove the tube from bicycle tires. by using handy tire levers, expert cyclist shows how to remove the tube from bicycle tires, when changing a flat tire, in this free bicycle repair video."

FP Examples:

\item "learn about the parts of a microscope with expert tips and advice on refurbishing antiques from a professional photographer in this free video clip on home astronomy and buildings. learn more about how to use a light microscope with a demonstration from a science teacher. "

\item "watch as a seasoned professional demonstrates how to use a deep fat fryer in this free online video about home pool care. get professional tips and advice from an expert on how to organize your kitchen appliance and kitchen appliance for special occasions."

\end{itemize}

\end{document}